\documentclass[conference]{IEEEtran}
\IEEEoverridecommandlockouts
\usepackage{cite}
\usepackage{amsmath,amssymb,amsfonts}
\usepackage{algorithmic}
\usepackage{graphicx}
\usepackage{textcomp}
\usepackage{xcolor}
\usepackage{soul}
\usepackage{url}
\usepackage[utf8]{inputenc}
\usepackage{graphicx}
\usepackage{amsmath}
\usepackage{booktabs}
\usepackage{algorithm}
\usepackage{algorithmic}
\usepackage{bbm}
\usepackage{listings}
\usepackage{xcolor}
\usepackage{multirow}
\newcommand{\tabincell}[2]{\begin{tabular}{@{}#1@{}}#2\end{tabular}}
\usepackage{stfloats}
\usepackage{float}
\usepackage{subfigure}
\usepackage{subcaption}
\usepackage{xspace}
\usepackage{tikz}
\usepackage{colortbl}
\usepackage{threeparttable}
\definecolor{mygray}{gray}{0.9}

\def\BibTeX{{\rm B\kern-.05em{\sc i\kern-.025em b}\kern-.08em
    T\kern-.1667em\lower.7ex\hbox{E}\kern-.125emX}}

\begin{document}

\title{ChatGraph: Interpretable Text Classification by Converting
ChatGPT Knowledge to Graphs \thanks{{$^\ast$} These authors contributed equally to this paper.}}

\author{\IEEEauthorblockN{1\textsuperscript{st} Yucheng Shi{$^\ast$}}
\IEEEauthorblockA{
\textit{University of Georgia}\\
Athens, GA \\
yucheng.shi@uga.edu}
\and
\IEEEauthorblockN{2\textsuperscript{nd} Hehuan Ma{$^\ast$}}
\IEEEauthorblockA{
\textit{University of Texas at Arlington}\\
Arlington, TX \\
hehuan.ma@mavs.uta.edu}
\and
\IEEEauthorblockN{3\textsuperscript{rd} Wenliang Zhong{$^\ast$}}
\IEEEauthorblockA{
\textit{University of Texas at Arlington}\\
Arlington, TX \\
wenliang.zhong@uta.edu}
\and
\IEEEauthorblockN{4\textsuperscript{th} Qiaoyu Tan}
\IEEEauthorblockA{
\textit{New York University Shanghai}\\
Shanghai, China \\
qiaoyu.tan@nyu.edu}
\and
\IEEEauthorblockN{5\textsuperscript{th} Gengchen Mai}
\IEEEauthorblockA{
\textit{University of Georgia}\\
Athens, GA \\
gengchen.mai25@uga.edu}
\and
\IEEEauthorblockN{6\textsuperscript{th} Xiang Li}
\IEEEauthorblockA{
\textit{Massachusetts General Hospital}\\
\textit{and Harvard Medical School}\\
Boston, Massachusetts \\
xli60@mgh.harvard.edu}
\and
\IEEEauthorblockN{7\textsuperscript{th} Tianming Liu}
\IEEEauthorblockA{
\textit{University of Georgia}\\
Athens, GA \\
tliu@uga.edu} 
\and
\IEEEauthorblockN{8\textsuperscript{th} Junzhou Huang}
\IEEEauthorblockA{
\textit{University of Texas at Arlington}\\
Arlington, TX \\
jzhuang@uta.edu}\\
}

\maketitle

\begin{abstract}
ChatGPT, as a recently launched large language model (LLM), has shown superior performance in various natural language processing (NLP) tasks.  However, two major limitations hinder its potential applications: 1) the inflexibility of finetuning on downstream tasks, and 2) the lack of interpretability in the decision-making process. To tackle these limitations, we propose a novel framework that leverages the power of ChatGPT for specific tasks, such as text classification, while improving its interpretability. The proposed framework conducts a knowledge graph extraction task to extract refined and structural knowledge from the raw data using ChatGPT. The rich knowledge is then converted into a graph, which is further used to train an interpretable linear classifier to make predictions. To evaluate the effectiveness of our proposed method, we conduct experiments on four benchmark datasets. The results demonstrate that our method can significantly improve the prediction performance compared to directly utilizing ChatGPT for text classification tasks. Furthermore, our method provides a more transparent decision-making process compared with previous text classification methods.
 The code is available at \url{https://github.com/sycny/ChatGraph}.
\end{abstract}

\section{Introduction}

Throughout the evolution of large language models (LLMs)\cite{johnson2016supervised,tai2015improved,zhu2015long,vaswani2017attention,kenton2019bert,liu2019roberta,hedeberta,sanh2019distilbert}, the recently launched ChatGPT has attracted mass attention in the NLP community~\cite{radford2018improving,radford2019language,brown2020language,raffel2020exploring,lewis2020bart,thoppilan2022lamda}. 
In a range of NLP tasks, including question answering, dialogue, summarization, named entity recognition, and sentiment analysis, ChatGPT has demonstrated impressive performance, outperforming many models even in zero-shot settings~\cite{nov2023putting,liu2023deid,shen2023hugginggpt,qin2023chatgpt,peng2023towards}. 

\begin{figure}[t]
    \centering
    \includegraphics[width=0.45\textwidth]{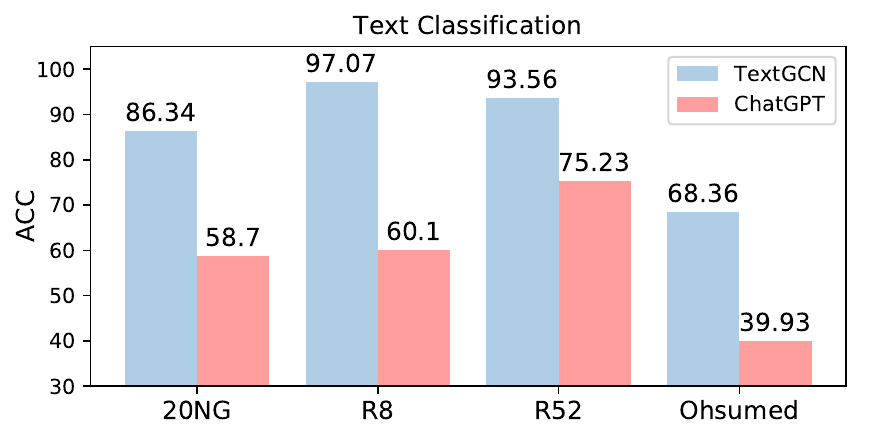}
    \caption{TextGCN (Blue) vs. ChatGPT (Red) in accuracy.}
    \label{intro_compare}
\end{figure}

However, despite the superior performance and generation ability, ChatGPT as a black-box model has two major limitations that hinder its potential applications. First, since model parameters are inaccessible, the ChatGPT model cannot be flexibly finetuned on specific datasets to adapt to certain tasks. Therefore, effective as it is, ChatGPT does not consistently outperform other models in every NLP task. For example, as illustrated in Figure~\ref{intro_compare}, TextGCN~\cite{yao2019graph} achieves significantly better performance than ChatGPT when used directly for text classification tasks. Second, similar to the traditional deep models, ChatGPT also suffers from the lack of transparency in the decision-making process, making it uninterpretable. Moreover, since ChatGPT is not open-sourced, existing interpretation methods such as attention scores cannot be applied. 



In this study, we tackle these two limitations in the text classification scenario. Specifically, to extend the potential of ChatGPT, it is necessary to consider two key questions. 
i) How to effectively utilize ChatGPT in scenarios where a substantial amount of labeled data is available, making it challenging to fully harness the potential of in-context learning~\cite{min2021recent}?
ii) How to enhance interpretability in the decision-making process? In this work, instead of developing a customized interpretation algorithm, we propose to tackle the problem from a data-centric perspective~\cite{zha2023data}. Specifically, given text data as input, rather than directly applying ChatGPT to classify text data, we first conduct a knowledge graph extraction task to extract refined and structural knowledge from the raw data using ChatGPT. The rich knowledge is then converted and distilled into a graph, which is further used to train an interpretable linear classifier to make predictions. Unlike traditional graph extraction methods, which mainly rely on heuristics~\cite{yao2019graph} or graph refinement~\cite{zhao2021data}, our proposed method does not require further refinement of the obtained graph, which reduces computational costs. Furthermore, our approaches provide a more transparent presentation of the relationships within context, and enable further investigation of the interpretation, thereby addressing the black box issue commonly associated with complex models. Additionally, our proposed framework is flexible and can be easily extended by integrating external knowledge to further enhance the training process, e.g., including TF-IDF weight~\cite{ramos2003using}.

Our contributions can be summarized from three perspectives. 1) We propose a novel framework that converts the learned knowledge from ChatGPT into a structural form, i.e., graphs, which enables more effective training of the knowledge with label information for text classification tasks. 2) The decision-making in our text classification framework is inherently interpretable, which overcomes the deficiency of black-box models in interpretability. 3) Our proposed method significantly improves the performance of ChatGPT over directly utilizing it for text classification tasks, which demonstrates the effectiveness and superiority of our approach.    

\begin{figure*}[t]
     \centering
     \includegraphics[width=0.97\textwidth]{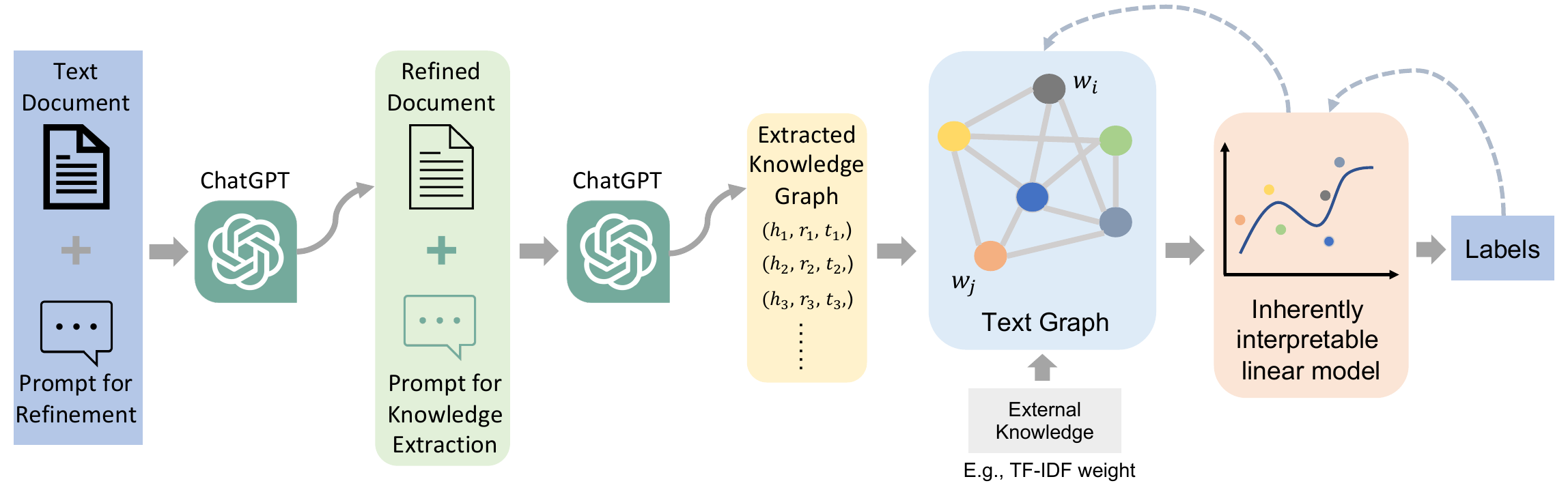}
\caption{ChatGraph framework for interpretable text classification: 1) raw text document is refined using ChatGPT with the designed prompt; 2) the knowledge graph is extracted from the refined document with another specific defined prompt; 3) the extracted knowledge graph is converted to text graph, where external knowledge can be inserted to assist the training; and 4) a linear model is trained on text graph for classification. }
    \label{framework}
 \end{figure*} 

\section{Methodology}
\subsection{Knowledge Graph Extraction with ChatGPT}
In the initial stage, we employ ChatGPT as a knowledge graph extractor to retrieve the semantic information from the raw text corpus. The extraction is established in two steps: 1) ChatGPT is employed to perform text refinement over the raw text to improve the general data quality; 2) the knowledge graph is extracted from the refined text utilizing ChatGPT with another adequate prompt.

\subsubsection{Text Refinement with ChatGPT}
As we noticed that the raw text includes certain deficiencies (i.e., typos and grammar mistakes), we propose to use ChatGPT to improve the input text quality. Specifically, we apply it to correct grammar and spelling errors, replace synonyms, and clarify the sentence structure of the original text.
This text refinement step helps ensure the input text is accurate and coherent, which is critical for the subsequent knowledge graph extraction. Formally, we denote the refined text corpus as $\mathcal{W}$, which contains a set of text segments $\mathcal{W} = \{W_1,...,W_{|\mathcal{W}|}\}$. And each text segment includes a sequence of words $W=(w_1,...,w_{|W|})$.

To enhance ChatGPT's text refinement abilities, we carefully develop task-specific prompts to activate its comprehension capabilities. The prompt is designed by incorporating human-like inquiries for better understanding, which is demonstrated below. 

\begin{lstlisting}
Please generate a refined document of the following document. And please ensure that the refined document meets the following criteria:
1. The refined document should be abstract and does not change any original meaning of the document. 
2. The refined document should retain all the important objects, concepts, and relationships between them. 
3. The refined document should only contain information that is from the document.
4. The refined document should be readable and easy to understand without any abbreviations and misspellings. 
Here is the content: [x]
\end{lstlisting}
\texttt{[x]} represents the placeholder for the raw text. As illustrated, our prompt seeks an abstract text that retains key information from the original document.

\subsubsection{Knowledge Graph Extraction from Refined Text} 
After refining the text, ChatGPT is utilized once again to extract the knowledge graph from the refined text. A knowledge graph is a set of entities and relations, where entities represent real-world objects or abstract concepts, and relations represent the relationships between entities~\cite{ji2021survey}. Knowledge graph extraction involves identifying entities and relationships from the text and representing them in a structured format, i.e., a triplet. Formally, we denote the knowledge graph as $\mathcal{KG}=\{\mathcal{T}, \mathcal{R}, \mathcal{F}\}$, where $\mathcal{T}$, $ \mathcal{R}$ and  $\mathcal{F}$ are sets of entities, relations, and facts, respectively. A fact can be represented as a triple $(h, r, t) \in \mathcal{F}$, which indicates the existence of a relation $r \in \mathcal{R}$ between the head entity $h \in \mathcal{T}$ and the tail entity $t \in \mathcal{T}$.

To facilitate knowledge graph extraction in ChatGPT, we adopt the concept of a chain of thoughts~\cite{wei2022chain} and design the prompt with step-by-step instructions. The initial prompts focus on identifying entities and relations, while the final prompts generate the desired output. The used prompt is shown as follows.
 \begin{lstlisting}
You are a knowledge graph extractor, and your task is to extract and return a knowledge graph from a given text.Let's extract it step by step:
(1). Identify the entities in the text. An entity can be a noun or a noun phrase that refers to a real-world object or an abstract concept. You can use a named entity recognition (NER) tool or a part-of-speech (POS) tagger to identify the entities.
(2). Identify the relationships between the entities. A relationship can be a verb or a prepositional phrase that connects two entities. You can use dependency parsing to identify the relationships.
(3). Summarize each entity and relation as short as possible and remove any stop words.
(4). Only return the knowledge graph in the triplet format: ('head entity', 'relation', 'tail entity').
(5). Most importantly, if you cannot find any knowledge, please just output: "None".
Here is the content: [x]
\end{lstlisting}

\subsection{Converting Knowledge Graph to Text Graph} 
After obtaining the knowledge graph $\mathcal{KG}$ from the refined document $\mathcal{W}$, we further build a graph with this information, namely text graph $\mathcal{G}$, to implement the text classification task. In such a manner, text graph $\mathcal{G}$ contains distilled semantic information extracted from the original text, which can be further used as the input of graph neural networks for classification. $\mathcal{G}$ is constructed by considering words as nodes and relational information as edges, which is formulated as $\mathcal{G}= \{\mathcal{V},\mathcal{E},\textbf{X},\textbf{A}\}$. $\mathcal{V}$ represents the set of nodes, which is formed with every distinct word appearing in the entities set $\mathcal{T}$ and relation set $\mathcal{R}$. $\textbf{X}\in \mathbbm{R}^{|{\mathcal{V}}|\times |\mathcal{V}|}$ denotes the feature matrix. For each node, we assign a one-hot vector as its attribute, where $\textbf{X} = \textbf{I}$, where $\textbf{I}$ is an identity matrix. $\mathcal{E}$ denotes the set of edges. An edge connection $e_{ij}$ is defined by observing two word nodes $v_i, v_j \in \mathcal{V}$ both appear in a fact triplet $(h,r,t)$. $\textbf{A} \in \left\{0,1\right\}^{|\mathcal{V}|\times |\mathcal{V}|}$ is the adjacency matrix, where $A_{ij}=1$ and $A_{ji}=1$ when edge $e_{ij}$ exists. 

In our proposed text graph, we explicitly save the hidden semantic relationships among different words in a structural format. 
Such transparency significantly enhances the interpretability of the text classification task. By utilizing the proposed text graph, we eliminate the need for black-box LLMs as classifiers, and can instead exploit simple linear models.

\subsection{Text Graph based Text Classification}
We adopt Graph Convolution Networks (GCN)~\cite{kipf2016semi} as the backbone of our classifier. To make it inherently interpretable, we only utilize one layer of GCN for the classification, as shown below:
\begin{equation}
\hat{\mathbf{y}}=\mathrm{Softmax}\left(\mathrm{Pooling}(\tilde{\mathbf{A}} \mathbf{X}\mathbf{W})\right),
\end{equation}
where $\hat{\mathbf{y}} \in \mathbbm{R}^{|\mathcal{W}| \times n}$ is the predicted label for each text segment, and $n$ is the number of label types. $\mathbf{W} \in \mathbbm{R}^{|\mathcal{V}| \times n}$ is a learnable weights matrix. The matrix $\tilde{\mathbf{A}}=\mathbf{D}^{-\frac{1}{2}} \bar{\mathbf{A}} \mathbf{D}^{-\frac{1}{2}}$ is used to normalize the adjacency matrix $\mathbf{A}$, where $\bar{\mathbf{A}}$ is the adjacency matrix with self-connections and $\mathbf{D}$ is the degree matrix and we have $\mathbf{D}_{ii} = {\textstyle \sum_{j}}\bar{\mathbf{A}}_{ij}$. In this study, we define the $\mathrm{Pooling}()$ operation as $\mathrm{Pooling}(\tilde{\mathbf{A}} \mathbf{X}\mathbf{W}) = \mathbf{S} \cdot (\tilde{\mathbf{A}} \mathbf{X}\mathbf{W})$. Here, $\mathbf{S} \in \{0,1\}^{|\mathcal{W}|\times |\mathcal{V}|}$ is the pooling matrix, which records the correspondence between each word and each text segment. If the corresponding word of node $v_i$ appears in text segments $W_j$, then $\mathbf{S}_{ji}=1$, else $\mathbf{S}_{ji}=0$. We choose the cross-entropy loss as our learning objective for the classification task, as shown below:
\begin{equation}
\mathcal{L}=-\sum_{i=1}^{|\mathcal{W}|} \mathbf{y}_i*\log(\hat{\mathbf{y}}[i]),
\end{equation}
where $\mathbf{y}_i$ is the ground truth label of text segment $W_i$. It is noteworthy that our classifier is inherently interpretable since it is a linear model without any non-linear activation functions.

\begin{table*}[htbp]
  \caption{Text classification performance comparison regarding accuracy scores (higher is better).}
  \label{text_classification}
  \centering
  \renewcommand\arraystretch{1.2}
\resizebox{0.85\textwidth}{!}{
\begin{threeparttable}
\begin{tabular}{cc|cccc}
\hline
Method   &Training Data       & 20NG                  & R8              & R52         & Ohsumed          \\
\hline \hline
 TF-IDF+LR & Full data & 83.19$_{\pm±0.00}$    & 93.74$_{\pm±0.00}$    & 86.95$_{\pm±0.00}$     & 54.66$_{\pm±0.00}$      \\
\hline
\tabincell{c}{TextGCN \\ (1 layer)}&Full data &78.85$_{\pm±0.10}$                   &  86.74$_{\pm±0.10}$              & 73.86$_{\pm±0.11}$         &50.25$_{\pm±0.08}$                           \\ 
\hline
 \tabincell{c}{TextGCN \\ (2 layers)}&Full data  &86.34$_{\pm±0.09}$                   &  97.07$_{\pm±0.10}$              & 93.56$_{\pm±0.18}$         &68.36$_{\pm±0.56}$                          \\
 \hline
\multirow{3}{*}{ChatGPT} & 0-shot & 58.70$_{\pm±0.00}$ & 60.10$_{\pm±0.00}$ & 75.23$_{\pm±0.00}$ & 39.93$_{\pm±0.00}$  \\
 & 2-shot & 58.44$_{\pm±0.00}$ & 72.54$_{\pm±0.00}$ & 81.68$_{\pm±0.00}$ & 47.05$_{\pm±0.00}$   \\
 & 5-shot & -- \tnote{1} & 82.43$_{\pm±0.00}$ & 90.13$_{\pm±0.00}$ & 45.39$_{\pm±0.00}$  \\
\hline
\rowcolor{mygray}ChatGraph &Full data  &79.15$_{\pm±0.08}$                   &  96.39$_{\pm±0.34}$       &92.14$_{\pm±0.26}$       &60.79$_{\pm±0.14}$                                   \\
\rowcolor{mygray}\tabincell{c}{ChatGraph \\ (with TF-IDF)} &Full data  &79.68 $_{\pm0.37}$                   & 96.46${\pm0.31}$ &93.25${\pm0.32}$ &63.63${\pm0.33}$      \\ 
\bottomrule
\end{tabular}
\begin{tablenotes}
    \scriptsize \item[1] Many samples in the 20NG dataset contain long sequences. Appending them in the prompt can easily exceed the maximum length allowed using the OpenAI API. We will leave the 5-shot text classification using ChatGPT for future work.
\end{tablenotes}
\end{threeparttable}
}
\end{table*}

\subsection{Integration with External Knowledge}
Our proposed text graph $\mathcal{G}$ can be easily extended by integrating various external knowledge to improve classification performance. In this section, we demonstrate how TF-IDF weight can be used as an example of external knowledge to enhance the pooling matrix.
 
Specifically, we replace the original weight in the pooling matrix with TF-IDF weight~\cite{ramos2003using}. The assumption of TF-IDF is that, if a word frequently appears in a segment, then it should be considered more critical to this segment. On the other hand, if the same word appears in many different text segments, then the word should be less important to that segment.
Formally, the TF-IDF score for a word $w$ in a text segment $W$ is calculated as:
\begin{equation}
    \text{tf-idf}(w,W) = \text{tf}(w,W) \times \text{idf}(w),
\end{equation}
where $\text{tf}(w,W)$ is the number of times for the word $w$ appears in a text segment $W$, which is calculated as:
\begin{equation*}
    \text{tf}(w,W) = \frac{\text{count}(w,W)}{|W|}.
\end{equation*}
The term $\text{count}(w,d)$ is the number of times for the word $w$ appears in text segment $W$, and $|W|$ is the total number of words in the document $W$.
$\text{idf}(w)$ is the inverse document frequency of the word $w$ in the corpus, which is defined as:
\begin{equation}
    \text{idf}(w) = \log\frac{|\mathcal{W}|}{df(w)},
\end{equation}
where $|\mathcal{W}|$ is the total number of text segments in the corpus, and $df(w)$ is the number of text segments in the corpus that contain the word $w$. 
The illustration of our proposed framework is demonstrated in Figure~\ref{framework}.

\section{Experiments}


\subsection{Experiment Settings}
\paragraph{Dataset Description}
We evaluate our proposed method on four commonly used text classification benchmark corpora, namely 20-Newsgroups (20NG), R8, R52 of Reuters 21578, and Ohsumed, which have been widely used in the field of text classification, and are considered to be representative of various types of text data~\cite{yao2019graph}. 
\begin{itemize}
    \item \textbf{20NG}, also known as 20 Newsgroups data, contains 18,846 newsgroup documents that are labeled with 20 categories with approx. equal sizes. In the experiments, 11,314 documents are utilized to train the model and the remaining 7,532 documents are used for testing. 
    \item \textbf{R8} and \textbf{R52} are two subsets that come from the Reuters-21578 dataset. R8 is categorized into 8 classes with 5,485 training and 2,189 testing documents. Similarly, R52 includes 52 classes with 6,532 documents used for training and 2,568 used for testing.
    \item \textbf{Ohsumed} is a collection of medical abstracts from MEDLINE, which is a medical literature dataset maintained by the National Library of Medicine. 13,929 documents with unique cardiovascular diseases are selected from the first 20,000 abstracts in the year 1991, e.g., neoplasms and eye diseases. Furthermore, documents with more than one label are excluded to form single-label classification, which results in a 7,400 dataset with 3,357 training data and 4,043 testing data.  
\end{itemize}

\paragraph{Baselines}
We conduct comparison experiments of ChatGraph with interpretable methods, graph-based methods, and ChatGPT with in-context learning to evaluate its effectiveness.  
\begin{itemize}
    \item \textbf{TF-IDF+LR} Firstly, we choose TF-IDF+LR\cite{yao2019graph} as our interpretable baseline, where a linear classifier is trained on the bag-of-words model with TF-IDF weight. 
    \item \textbf{TextGCN} Then we select TextGCN\cite{yao2019graph} as the graph-based baseline, where a graph is constructed based on the vocabulary of words and documents. It leverages GCNs to tackle text classification as a node classification problem. We explore two variations of TextGCN: the first utilizes a single GCN layer and is considered inherently interpretable, while the second follows the original architecture with two GCN layers while lacking interpretability, resembling a black-box model.
    \item \textbf{ChatGPT} Finally, we examine large language models (LLMs) that leverage prompts. Recent research~\cite{brown2020language} has highlighted the power of LLMs in zero/few-shot learning through prompt engineering without extensive training. For our experiments, we design a $k$-shot prompt specifically for ChatGPT to perform classification. The prompt instructs ChatGPT to categorize a given text document into predefined classes and provides $k$ examples for the selected class. In cases where ChatGPT is unable to determine the classification, we simply require it to output "None."
\end{itemize}
The implementation details for $k$-shot ChatGPT can be found in the code. The example prompt for R8 datasets is included in the Appendix.

\begin{table*}[!htbp]
\tiny
\footnotesize
\caption{Case study of test examples. (Important words are highlighted in red and blue).} \label{tb3}
\centering
\resizebox{\textwidth}{!}{
\begin{tabular}{cl}

\hline \hline
\multicolumn{2}{l}{\emph{Case 1:} Test Sample From \textit{R8}, label = ``interest"} \\
\hline
Raw Text: &  bank of france leaves intervention rate unchanged at pct official. \\
Refined Text: & The \textcolor[RGB]{0, 0, 255}{\textbf{Bank}} of \textcolor[RGB]{0, 0, 255}{\textbf{France}} has decided to \textcolor[RGB]{0, 0, 255}{\textbf{maintain}} its intervention \textcolor[RGB]{255, 0, 0}{\textbf{rate}} at the current percentage, according to an \textcolor[RGB]{0, 0, 255}{\textbf{official}} statement. \\
Extracted KG:& (`Bank of France', `maintain', `intervention rate') \\

\hline\hline
\multicolumn{2}{l}{\emph{Case 2:} Test Sample From \textit{R52}, label = ``earn"} \\
\hline
Raw Text: & kiena two for one share split approved kiena gold mines ltd said shareholders approved a previously reported proposed two for one common \\& stock split record date of the split will be april kiena said reuter \\
Refined Text: 
& Kiena Gold Mines Ltd has announced that its \textcolor[RGB]{255, 0, 0}{\textbf{shareholders}} have approved a proposed \textcolor[RGB]{0, 0, 255}{\textbf{two-for-one}} common \textcolor[RGB]{0, 0, 255}{\textbf{stock split}}. The \textcolor[RGB]{0, 0, 255}{\textbf{record}} date \\& for the split has been set for April.\\
Extracted KG: 
&(`Kiena Gold Mines Ltd', `announced', `shareholders') (`shareholders', `approved', `proposed two-for-one common stock split') (`record date', `set for', `April')\\

\hline\hline
\multicolumn{2}{l}{\emph{Case 3:} Test Sample From \textit{R52}, label = ``coffee"} \\
\hline
Raw Text: &coffee could drop to cts cardenas says international coffee prices could drop to between and cents a lb by next october if no agreement is \\ & reached to support the market jorge cardenas manager of colombia s national coffee growers federation said speaking at a forum for industrialists  \\ & he said one of the reasons was that the market was  already saturated and that producers will have excess production  and stockpiles of mln kg bags \\ & in today may futures in new york settled at cents a lb reuter   \\
Refined Text: &International \textcolor[RGB]{255, 0, 0}{\textbf{coffee}} prices may decrease to a range of 90 to 100 cents per pound by October of next year if no measures are taken to \\ &  boost the market, according to \textcolor[RGB]{0, 0, 255}{\textbf{Jorge Cardenas}}, the manager of \textcolor[RGB]{0, 0, 255}{\textbf{Colombia's}} National \textcolor[RGB]{255, 0, 0}{\textbf{Coffee}} Growers Federation. Speaking at an industrial forum, \\ & Cardenas cited market saturation and excess production as reasons for the potential price drop. Currently, there are stockpiles of 1 million \textcolor[RGB]{0, 0, 255}{\textbf{kg bags}}. \\ & As of today, May futures in New York settled at 106 cents per pound. \\
Extracted KG:&(`International coffee prices', `may decrease to', `a range of 90 to 100 cents per pound')\\&(`measures', `are taken to boost', `the market') \\ &
(`Jorge Cardenas', `is', `the manager of Colombia\'s National Coffee Growers Federation')\\ &(`Cardenas', `cited', `market saturation and excess production as reasons for the potential price drop')\\ &(`there', `are', `stockpiles of 1 million kg bags')\\&(`May futures in New York', `settled at', `106 cents per pound') \\
\hline \hline

\end{tabular}
}
\end{table*}

\subsection{Experiment Results}
The comparison results of text classification accuracy are shown in Table~\ref{text_classification}. As observed, 1) TF-IDF with logistic regression performs well and is interpretable. However, TextGCN with two GCN layers outperforms TF-IDF, indicating its ability to capture complex patterns; 2) for TextGCN model, there is a tradeoff between performance and interpretability with different layer architectures. Deeper models lack interpretability, while one-layer models have lower performance; 3) ChatGPT's zero-shot performance is poor but improves by including a few labeled examples. Nevertheless, the performance still lags behind TextGCN. Additionally, ChatGPT struggles with long input texts like 20NG due to token length limitations; 4) our proposed method, ChatGraph, surpasses other inherently explainable baselines and outperforms ChatGPT in zero/few-shot scenarios. Additionally, when combined with TF-IDF, ChatGraph's performance is further enhanced. These findings indicate that our approach achieves a balance between performance and interpretability, making it a promising solution for text classification tasks that demand both high performance and interpretability.

\subsection{Evaluation on Interpretability}

In this section, we evaluate the interpretability of our proposed model. Since we use a naturally interpretable linear model for text classification, the weights of $\mathbf{W}$ in the linear model correspond to the importance of each word in determining the label, providing a straightforward explanation of the model's decision-making process. To quantitatively demonstrate the interpretability of our approach, we compare fidelity as interpretability metrics:
$
    Fid_{m} =\frac{1}{N} \sum_{i=1}^{N}\left(\mathbbm{1}\left(\hat{\mathbf{y}}_{i}=\mathbf{y}_{i}\right)-\mathbbm{1}\left(\hat{\mathbf{y}}_{i}^{m}=\mathbf{y}_{i}\right)\right) \times 100\%,
$
where $\hat{\mathbf{y}}_{i}^{m}$ denotes the prediction with $m$ percent features are masked. In Table~\ref{fid}, we present the fidelity scores with 0.1\% and 1\% of the features masked. We compare two masking strategies: Rand mask (features are randomly masked), and Top mask (masking is applied to the most important features). We can observe that masking features with high explanation values have higher fidelity scores, which empirically indicates our interpretation is accurate.
\begin{table}[!th]
\small
  \renewcommand\arraystretch{1.3}
  \caption{Fidelity metrics of our proposed methods. (A higher value indicates important features are masked.)}
  \label{fid}
  \centering
\scalebox{1}{
\begin{tabular}{c|cccc}
\hline
Mask ratio          & 20NG                  & R8              & R52         & Ohsumed          \\\hline\hline
Rand mask-0.1\%    & 2.04            & 1.41                        & 0.31  &0.05  \\
Top mask-0.1\%      & \textbf{68.00}             & \textbf{24.44}                        & \textbf{52.33}     & \textbf{54.43}     \\ \hline
Rand mask-1\%     &1.29              & 1.32                          & 0.35  &0.25  \\ 
Top mask-1\%  &\textbf{70.23}              & \textbf{38.46}                       &\textbf{85.55}   &\textbf{56.83}  \\ \hline
\end{tabular}
}
\end{table}

To qualitatively demonstrate the interpretability of our approach, we selected three samples from the R8 and R52 datasets for sample analysis. The results of this analysis are presented in Table~\ref{tb3}, which shows the original text, the refined text, and the extracted knowledge graph. We have highlighted the words that are most important for text classification in red, and the next four most important words in blue. Upon reviewing the results of our model, it is evident that our approach is capable of accurately identifying the words that are most critical in determining the label.

\section{Conclusion}
We introduce a novel and interpretable text classification framework leveraging ChatGPT. Extensive experiments demonstrate that the proposed ChatGraph not only improves the text classification performance with inherent interpretability, but also provides innovative insights for exploring the application of LLM models. 
In our next step, we will extend our method to other NLP tasks, and further explore the interpretability of ChatGPT.



\bibliographystyle{IEEEtran}
\bibliography{ICDM}

\appendix
\section{Example Prompt for ChatGPT}
\label{ChatGPTprompt}
We include the prompt we designed for the ChatGPT on the text classification task as a reference.

\begin{lstlisting}
You are a text classifier and your task is to classifiy a given text into the following categories: ['acq', 'crude', 'earn', 'grain', 'interest', 'money-fx', 'ship', 'trade']. You should directly output the predicted label only. You answer should be either one of ['acq', 'crude', 'earn', 'grain', 'interest', 'money-fx', 'ship', 'trade']. Do not output a sentence.

Good example: 
###Input###:
champion products approves stock split champion products inc said board directors approved two one stock split common shares shareholders record april company also said board voted recommend shareholders annual meeting april increase authorized capital stock five mln mln shares reuter. 
###Output###:
earn

Bad example:
###Input###:
champion products approves stock split champion products inc said board directors approved two one stock split common shares shareholders record april company also said board voted recommend shareholders annual meeting april increase authorized capital stock five mln mln shares reuter. 
###Output###:
loss
\end{lstlisting}

\end{document}